\documentclass[10pt,twocolumn,letterpaper]{article}

\usepackage[pagenumbers]{cvpr}

\usepackage{graphicx}
\usepackage{grffile}
\usepackage{capt-of}
\usepackage{xcolor}
\usepackage{amsmath,amssymb,amsfonts,dsfont,pifont,bm,bbm,mathrsfs,mathtools,nicefrac}
\usepackage{algorithm,algpseudocode,listings}
\usepackage{booktabs,multirow,adjustbox,diagbox,threeparttable}
\usepackage{subcaption}
\definecolor{citeblue}{RGB}{48,111,186}
\usepackage[pagebackref,breaklinks,colorlinks,citecolor=citeblue,bookmarks=false]{hyperref}
\usepackage[capitalize]{cleveref}
\crefname{section}{Sec.}{Secs.}
\Crefname{section}{Section}{Sections}
\crefname{table}{Tab.}{Tabs.}
\Crefname{table}{Table}{Tables}
\crefname{figure}{Fig.}{Figs.}
\Crefname{figure}{Figure}{Figures}
\crefname{equation}{Eq.}{Eqs.}
\Crefname{equation}{Equation}{Equations}
\usepackage{color, colortbl}
\definecolor{LightGray}{gray}{0.9}

\usepackage{multirow}
\usepackage{array}
\newcolumntype{L}[1]{>{\raggedright\let\newline\\\arraybackslash\hspace{0pt}}m{#1}}
\newcolumntype{C}[1]{>{\centering\let\newline\\\arraybackslash\hspace{0pt}}m{#1}}
\newcolumntype{R}[1]{>{\raggedleft\let\newline\\\arraybackslash\hspace{0pt}}m{#1}}

\newcommand{\cmark}{\ding{51}}
\newcommand{\xmark}{\ding{55}}

\newcommand{\method}{\textit{DisCoScene}\xspace}
\newcommand{\supp}{\textit{Supplementary Materials}\xspace}
\newcommand{\clevr}{\textsc{Clevr}\xspace}
\newcommand{\front}{\textsc{3D-Front}\xspace}
\newcommand{\waymo}{\textsc{Waymo}\xspace}

\newcommand{\tocite}[1]{\textcolor{red}{[TO CITE]}}

\definecolor{color1}{HTML}{548235}
\definecolor{color2}{HTML}{0070C0}

\def\cvprPaperID{170}
\def\confName{CVPR}
\def\confYear{2023}

\usepackage{fp}
\usepackage{expl3}[2012-07-08]
\ExplSyntaxOn
\ExplSyntaxOff

\usepackage{amsmath,amsfonts,bm}

\newcommand{\ignorethis}[1]{}

\def\eqref#1{equation~\ref{#1}}

\def\1{\bm{1}}

\def\va{{\bm{a}}}

\def\vc{{\bm{c}}}

\def\vs{{\bm{s}}}
\def\vt{{\bm{t}}}

\def\vv{{\bm{v}}}
\def\vw{{\bm{w}}}
\def\vx{{\bm{x}}}

\def\vz{{\bm{z}}}

\DeclareMathAlphabet{\mathsfit}{\encodingdefault}{\sfdefault}{m}{sl}
\SetMathAlphabet{\mathsfit}{bold}{\encodingdefault}{\sfdefault}{bx}{n}

\newcommand{\ignore}[1]{}

\begin{document}

\title{DisCoScene: Spatially Disentangled Generative Radiance Fields \\ for Controllable 3D-aware Scene Synthesis}

\author{Yinghao Xu$^{1,2}$\thanks{Work partially done during internships at Snap Inc.} \quad Menglei Chai$^{2}$\thanks{Corresponding author.} \quad Zifan Shi$^{3}$ \quad Sida Peng$^{4}$ \\ Ivan Skorokhodov$^{5,2*}$ \quad Aliaksandr Siarohin$^{2}$ \quad Ceyuan Yang$^{1}$ \quad Yujun Shen$^{1}$ \\ Hsin-Ying Lee$^{2}$ \quad Bolei Zhou$^{6}$ \quad Sergey Tulyakov$^{2}$ \\[5pt]
	{$^1$CUHK \quad
	$^2$Snap Inc. \quad
	$^3$HKUST \quad 
    $^4$ZJU \quad 
    $^5$KAUST \quad 
    $^6$UCLA}
    \\
	}

\maketitle

\begin{abstract}
    Existing 3D-aware image synthesis approaches mainly focus on generating a single canonical object and show limited capacity in composing a complex scene containing a variety of objects.
    This work presents DisCoScene: a 3D-aware generative model for high-quality and controllable scene synthesis.
    The key ingredient of our method is a very abstract object-level representation (\ie, 3D bounding boxes without semantic annotation) as the scene layout prior, which is simple to obtain, general to describe various scene contents, and yet informative to disentangle objects and background.
    Moreover, it serves as an intuitive user control for scene editing.
    Based on such a prior, the proposed model spatially disentangles the whole scene into object-centric generative radiance fields by learning on only 2D images with the global-local discrimination.
    Our model obtains the generation fidelity and editing flexibility of individual objects while being able to efficiently compose objects and the background into a complete scene. 
    We demonstrate state-of-the-art performance on many scene datasets, including the challenging Waymo outdoor dataset.
    Project page can be found \href{https://snap-research.github.io/discoscene/}{here}.
\end{abstract}

\section{Introduction}\label{sec:intro}

3D-consistent image synthesis from single-view 2D data has become a trendy topic in generative modeling.
Recent approaches like GRAF\cite{graf} and Pi-GAN~\cite{pigan} introduce 3D inductive bias by taking neural radiance fields~\cite{nerf, deepsdf, occupancy, neuralbody, mipnerf} as the underlying representation, gaining the capability of geometry modeling and explicit camera control.
Despite their success in synthesizing individual objects (\eg, faces, cats, cars), they struggle on scene images that contain multiple objects with non-trivial layouts and complex backgrounds.
The varying quantity and large diversity of objects, along with the intricate spatial arrangement and mutual occlusions, bring enormous challenges, which exceed the capacity of the object-level generative models~\cite{ eg3d, volumegan, epigraf, stylesdf, stylenerf, gao2022get3d, geod}.

Recent efforts have been made towards 3D-aware scene synthesis.
Despite the encouraging progress, there are still fundamental drawbacks.
For example, Generative Scene Networks (GSN)~\cite{gsn} achieve large-scale scene synthesis by representing the scene as a grid of local radiance fields and training on 2D observations from continuous camera paths.
However, object-level editing is not feasible due to spatial entanglement and the lack of explicit object definition.
On the contrary, GIRAFFE~\cite{giraffe} explicitly composites object-centric radiance fields \cite{objectnerf, guo2020object, neuralscene, wu2022objectsdf} to support object-level control.
Yet, it works poorly on challenging datasets containing multiple objects and complex backgrounds due to the absence of proper spatial priors.

To achieve high-quality and controllable scene synthesis, the scene representation stands out as one critical design focus. A well-structured scene representation can scale up the generation capability and tackle the aforementioned challenges. Imagine, given an empty apartment and a furniture catalog, what does it take for a person to arrange the space? Would people prefer to walk around and throw things here and there, or instead figure out an overall layout and then attend to each location for the detailed selection? Obviously, a layout describing the arrangement of each furniture in the space substantially eases the scene composition process~\cite{li2019layoutgan, blockplanner, gupta2021layouttransformer}.
From this vantage point, here comes our primary motivation --- an abstract object-oriented scene representation, namely a \textit{layout prior}, could facilitate learning from challenging 2D data as a lightweight supervision signal during training and allow user interaction during inference.
More specifically, to make such a prior easy to obtain and generalizable across different scenes, we define it as a set of object bounding boxes without semantic annotation, which describes the spatial composition of objects in the scene and supports intuitive object-level editing. 

In this work, we present \textit{DisCoScene}, a novel 3D-aware generative model for complex scenes. Our method allows for high-quality scene synthesis on challenging datasets and flexible user control of both the camera and scene objects. Driven by the aforementioned \textit{layout prior}, our model spatially disentangles the scene into compositable radiance fields which are shared in the same object-centric generative model. To make the best use of the prior as a lightweight supervision during training, we propose global-local discrimination which attends to both the whole scene and individual objects to enforce spatial disentanglement between objects and against the background. Once the model is trained, users can generate and edit a scene by explicitly controlling the camera and the layout of objects' bounding boxes.
In addition, we develop an efficient rendering pipeline tailored for the spatially-disentangled radiance fields, which significantly accelerates object rendering and scene composition for both training and inference stages.

Our method is evaluated on diverse datasets, including both indoor and outdoor scenes. Qualitative and quantitative results demonstrate that, compared to existing baselines, our method achieves state-of-the-art performance in terms of both generation quality and editing capability. 
\cref{tab:contributions} compares \method with relevant works.
it is worth noting that, to the best of our knowledge, \method stands as the first method that achieves high-quality 3D-aware generation on challenging datasets like \waymo~\cite{waymo}, while enabling interactive object manipulation.

\section{Related Work}\label{sec:related-work}

\noindent{\textbf{3D-aware Image Synthesis}}. 
Generative Adversarial Networks (GANs) have achieved remarkable success in 2D image synthesis~\cite{gan, pggan, stylegan, stylegan2, stylegan3},
and have recently been extended to 3D-aware image generation.
VON\cite{von} and HoloGAN\cite{hologan} introduce voxel representations to the generator and use neural rendering to project 3D voxels into 2D space.
Then, GRAF\cite{graf} and Pi-GAN\cite{pigan} propose to use implicit functions to learn NeRF from single-view image collections, resulting in better multi-view consistency compared to voxel-based methods.
GOF~\cite{gof}, ShadeGAN~\cite{shadegan}, and GRAM~\cite{gram} introduce occupancy field, albedo field and radiance surface instead of radiance field to learn better 3D shapes.
However, high-resolution image synthesis with direct volumetric rendering is usually expensive.
Many works~\cite{giraffe, stylenerf, eg3d, volumegan, stylesdf, xue2022giraffehd, geod} resort to convolutional upsamplers to improve the rendering resolution and quality with lower computation overhead.
While some other works~\cite{epigraf, voxgraf} adopt patch-based sampling and sparse-voxel to speed up training and inference.
Note that most of these methods are restricted to well-aligned objects and fail on more complex, multi-object scene imagery.
Our work instead naturally handles multi-object scenes with spatial disentangled object-level radiance fields, which can be scaled to very challenging real-world scene datasets. 

\setlength{\tabcolsep}{0\linewidth}
\begin{table}[t]
\footnotesize
    \centering
    \caption{\textbf{Comparison of \method and relevant works.}
    \textit{Multiple Objects}: Ability to model multiple objects in a scene. 
    \textit{Radiance Field}: If radiance fields are used to model scenes. 
    \textit{Complex Scene}: Ability to handle complex datasets beyond diagnostic scenes.
    \textit{Object Editing}: If object-level editing is supported.
    \textit{No Camera Sequence}: Not requiring ground truth camera sequences.
    }
    \label{tab:contributions}
    \vspace{-8pt}
    \begin{tabular}{C{0.225\linewidth}C{0.155\linewidth}C{0.155\linewidth}C{0.155\linewidth}C{0.155\linewidth}C{0.155\linewidth}}
    \toprule
     \multirow{2}{*}{Model} &Multiple&Radiance& Complex&Object&No Camera\\
     &Objects&Field&Scene&Editing&Sequence\\
    \midrule 
    GRAF \cite{graf} & \xmark   & \cmark  & \xmark  & \cmark & \cmark\\
    BlockGAN \cite{hologan} & \cmark   & \xmark & \xmark & \cmark & \cmark\\ 
    GSN~\cite{gsn}  & \cmark  & \cmark  & \cmark & \xmark & \xmark\\ 
    GIRAFFE \cite{giraffe} & \cmark   & \cmark  & \xmark & \cmark & \cmark\\ 
    \rowcolor{LightGray}
    DisCoScene  & \cmark & \cmark  & \cmark & \cmark & \cmark\\ 
    \bottomrule
    \end{tabular}
    \vspace{-10pt}
\end{table}

\noindent{\textbf{Scene Generation}}. 
Scene generation has been a long-standing task.
Early works like~\cite{tu2005image} attempt to model a complex scene by trying to generate it.
Recently, with the successes in generative models, scene generation has been advanced significantly.
Among them, one popular line is to resort to the setups of image-to-image translation from given conditions, \ie, semantic masks~\cite{zhu2017unpaired, isola2017image, wang2018high, park2019semantic, tan2021efficient, tan2022semantic}, object-attribute graph~\cite{bear2020learning}.
Although able to synthesize photorealistic scene images, they struggle to manipulate the objects in 3D space due to the lack of 3D understanding.
Some works~\cite{yang2021semantic, wang2022improving, zhang2021decorating, blobgan} reuse the knowledge from 2D GAN models to achieve scene manipulation like the camera pose.
But they suffer from poor multi-view consistency due to inadequate geometry modeling.
Another active line of work~\cite{nguyen2020blockgan, giraffe, yang2021semantic, blobgan} explores adding 3D inductive biases to the scene representation.
BlockGAN~\cite{nguyen2020blockgan} and GIRAFFE~\cite{giraffe} introduce compositional voxels and radiance fields to encode the object structures, but their object control can only be performed at simple diagnostic scenes.
DepthGAN~\cite{depthgan} introduces depth as a 3D prior but is hard to achieve manipulation and multi-view consistency.
GSN~\cite{gsn} proposes to represent a scene with a grid of local radiance fields.
However, since this local radiance field does not properly link to the object semantics, individual objects cannot be manipulated with versatile user control.
Our work proposes to use an abstract layout prior to spatially disentangle the whole scene into object-centric radiance fields, which enables 3D-aware image synthesis on challenging real-world imagery like \waymo~\cite{waymo}.

\section{Method}\label{sec:method}

\begin{figure*}[t]
\begin{center}
\includegraphics[width=0.87\textwidth]{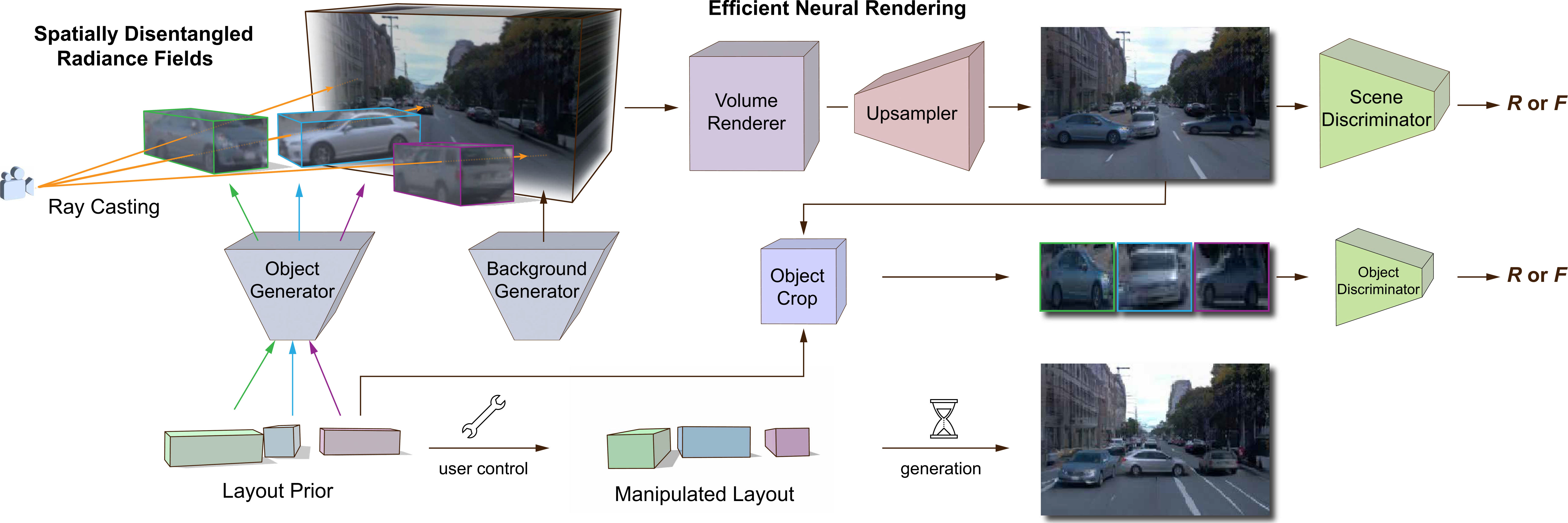}
\end{center}
\vspace{-16pt}
\caption{\textbf{The overall pipeline of \method.} Conditioned by the \textit{layout prior}, our \textit{spatial disentangle generative radiance fields} generate individual objects and the background. Our \textit{efficient neural rendering} pipeline then composites the scene to a low-resolution feature map with the \textit{volume renderer} and upsamples to the final high-resolution image with the \textit{upsampler}.
During training, we propose \textit{global-local discrimination} which applies the \textit{scene discriminator} to the entire image and the \textit{object discriminator} to cropped object patches.
During inference, users can manipulate the layout to control the generation of a specific scene at the object level.}
\label{fig:pipeline}
\vspace{-10pt}
\end{figure*}

The overall framework is illustrated in \cref{fig:pipeline}.
We employ layout as an explicit prior to disentangle objects in our approach (\cref{sec:method:layout}).
Based on the layout prior, we introduce our spatially disentangled radiance fields (\cref{sec:method:radiance-field}) and an efficient rendering pipeline (\cref{sec:method:rendering}) to achieve controllable 3D-aware scene generation.
We also describe our global-local discrimination, which makes training on challenging datasets possible (\cref{sec:method:discrimination}).
Finally, we discuss our model's training and inference details on 2D image collections (\cref{sec:method:training}).

\subsection{Abstract Layout Prior}\label{sec:method:layout}
There exist many representations of a scene, including the popular choice of scene graph~\cite{ImageFromSceneGraph, sowizral2000scene, neuralscene, cunningham2001lessons}, where objects and their relations are denoted as nodes and edges.
Although graph can describe a scene in rich details, its structure is hard to process and the annotation is laborious to obtain in our case.
Therefore, we opt to represent the scene layout in a much-simplified manner -- a set of bounding boxes $\mathcal{B} = \{\mathbf{B}_i| i \in [1, N]\}$ without category annotation, where $N$ counts objects in the scene.
Concretely, each bounding box is defined with $9$ parameters, including rotation $\va_i$, translation $\vt_i$, and scale $\vs_i$.
\begin{align}
    \mathbf{B}_i &= [\va_i, \vt_i, \vs_i], \\
    \va_i = [a_x, a_y, a_z], 
    \vt_i &= [t_x, t_y, t_z], 
    \vs_i = [s_x, s_y, s_z], 
\end{align}
where $\va_i$ comprises $3$ Euler angles, which are easier to convert into rotation matrix $\mathbf{R}_i$.
Using this notation, the bounding box $\mathbf{B}_i$ can be transformed from a canonical bounding box $\mathbf{C}$, \ie, a unit cube at the coordinate origin:
\begin{align}
    \mathbf{B}_i = \mathrm{b}_{i}(\mathbf{C}) =  \mathbf{R}_i \cdot \mathrm{diag}(\vs_i) \cdot \mathbf{C}  + \vt_i,
\end{align}
where $\mathrm{b}_{i}$ stands for the transformation of $\mathbf{B}_i$ and $\mathrm{diag}(\cdot)$ yields a diagonal matrix with the elements of $\vs_i$. 
Our abstract layout is more friendly to collect and easier to edit, allowing for versatile interactive user control.

\subsection{Spatially Disentangled Radiance Fields}\label{sec:method:radiance-field}

\noindent{\textbf{Object Representation}.}
Neural radiance field (NeRF)~\cite{nerf} $\mathrm{F}(\vx, \vv) \rightarrow (\vc, \sigma)$ regresses color $\vc \in \mathbb{R}^3$ and volume density $\sigma \in \mathbb{R}$ from coordinate $\vx \in \mathbb{R}^3$ and viewing direction $\vv \in \mathbb{S}^2$, parameterized with multi-layer perceptron (MLP) networks.
Recent attempts propose to condition NeRF with a latent code $\vz$, resulting in their generative forms~\cite{graf, pigan}, $\mathrm{G}(\vx, \vv, \vz) \rightarrow (\vc, \sigma)$, to achieve 3D-aware object synthesis.

Since we use the layout as an internal representation, it naturally disentangles the whole scene into several objects.
We can leverage multiple individual generative NeRFs to model different objects, but it can easily lead to an overwhelmingly large number of models and poor training efficiency. 
To alleviate this issue, we propose to infer generative object radiance field in the \textit{canonical} space, to allow weight sharing among objects:
\begin{align}
    (\vc_i, \sigma_i) = \mathrm{G_{obj}}(\mathrm{b}_{i}^{-1}(\gamma(\vx)), \vz_i), \label{eq:fg}
\end{align}
where $\gamma(\cdot)$ is the position encoding function that transforms input into Fourier features. The object generator $\mathrm{G_{obj}}(\cdot)$ infers each object independently, resulting in \textit{spatially disentangled generative radiance fields}. 
Note that $\mathrm{G_{obj}}(\cdot)$ is not conditioned on the viewing direction $\vv$ because the upsampler of our neural renderer can learn the view-dependent effects similar to previous works~\cite{stylenerf, volumegan, eg3d} (\cref{sec:method:rendering}).

\noindent{\textbf{Spatial Condition}.}
Although object bounding boxes are used as a prior, their latents are still randomly sampled regardless of their spatial configuration, leading to illogical arrangements.
To synthesize scene images and infer object radiance fields with proper semantics,
we adopt the location and scale of each object as a condition for the generator to encode more consistent intrinsic properties, \ie, shape and category.
To this end, we simply modify \cref{eq:fg} by concatenating the latent code with the Fourier features of object location and scale:
\begin{align}
    (\vc, \sigma) = \mathrm{G_{obj}}(\mathrm{b}_{i}^{-1}(\gamma(\vx)), \mathrm{concat}(\vz_i, \gamma(\vt_i), \gamma(\vs_i)). \label{eq:fg-new}
\end{align}
Therefore, semantic clues are injected into the layout in an unsupervised manner, without explicit category annotation.

\noindent{\textbf{Background Representation}.}
Unlike objects, the background radiance field is only evaluated in the global space.
Considering that the background encodes lots of high-frequency signals, we include the viewing direction $\vv$ to help background generator $\mathrm{G_{bg}}(\cdot)$ to be able to learn such details.
The background generation can be formulated as:
\begin{align}
    (\vc_\mathrm{bg}, \sigma_\mathrm{bg}) = \mathrm{G_{bg}}(\vx, \vv, \vz_\mathrm{bg}). \label{eq:bg}
\end{align}

\subsection{Efficient Rendering Pipeline} \label{sec:method:rendering}
As aforementioned, we use spatial-disentangled radiance fields to represent scenes. However, na\"ive point sampling solutions can lead to prohibitive computational overhead when rendering multiple radiance fields.
Considering the independence of objects' radiance fields, we can achieve much more efficient rendering by only focusing on the valid points within the bounding boxes.

\noindent{\textbf{Ray-Box Intersection in Canonical Space.}}
Similar to NeRF~\cite{nerf}, we use the pinhole camera model to perform ray casting.
For each object, the points on the rays can be sampled at adaptive depths rather than fixed ones since the bounding box provides clues about where the object locates.
Specifically, the cast rays $\mathcal{R} = \{\mathbf{r}_j | j\in[1, \mathrm{S}^2]\}$ in a resolution $\mathrm{S}$ are transformed into the canonical object coordinate system.
Then, Ray-AABB (axis-aligned bounding box) intersection algorithm is applied to calculate the adaptive near and far depth $(d_{j, l, n},d_{j, l, f})$ of the intersected segment between the ray $\mathbf{r}_j$ and the $l$-th box $\mathbf{B}_l$.
After that, $\mathrm{N_d}$ points are sampled equidistantly in the interval $[d_{j, l, n}, d_{j, l, f}]$.
It is worth noting that we maintain an intersection matrix $\mathbf{M}$ of size $\mathrm{N}\times \mathrm{S}^2$, whose elements indicate if this ray intersects with the box.
With $\mathbf{M}$, we select only valid points to infer, which can greatly reduce the rendering cost.

\noindent{\textbf{Background Point Sampling}}. 
We adopt different background sampling strategies depending on the dataset.
In general, we do fixed depth sampling for bounded backgrounds in indoor scenes and inherit the inverse parametrization of NeRF++~\cite{nerf++} for complex and unbounded outdoor scenes, which uniformly samples background points in an inverse depth range.
More details can be founded in the \textit{Supplementary Materials}.

\noindent{\textbf{Composition and Volume Rendering}.} 
In our approach, objects are always assumed to be in front of the background.
So objects and background can be rendered independently first and composited thereafter.
For a ray $\mathbf{r}_j$ intersecting with $n_j$ ($n_j \geq 1$) boxes, its sample points $\mathcal{X}_j = \{\vx_{j, k} | k \in [1, n_j\mathrm{N}_d] \}$ can be easily obtained from the depth range and the intersection matrix $\mathbf{M}$. 
Since rendering should consider inter-object occlusions, we sort the points $\mathcal{X}$ by depth, resulting in an ordered point set $\mathcal{X}_j^{s} = \{\vx_{j, s_k} | s_k \in [1, n_j\mathrm{N}_d], d_{j, s_k} \leq d_{j, s_{k+1}} \}$, where $d_{j, s_k}$ denotes the depth of point $\vx_{j, s_k}$.
With color $\vc(\vx_{j, s_k})$ and density $\sigma(\vx_{j, s_k})$ of the ordered set inferred with $\mathrm{G_{obj}(\cdot)}$ by \cref{eq:fg-new}, 
the corresponding pixel $\mathbf{f}(\mathbf{r}_j)$ is calculated as: 
\begin{align}
        \mathbf{f}(\mathbf{r}_j) &= \sum_{k=1}^{n_j\mathrm{N}_d} T_{j,k} \alpha_{j, k} \vc(\vx_{j, s_k}), \label{eq:integral}\\
    T_{j, k} &= \exp(-\sum_{o=1}^{k-1} \sigma(\vx_{j, s_k}) \delta_{j, s_o}), \\
    \alpha_{j, k} &= 1 - \exp(-\sigma(\vx_{j, s_k}) \delta_{j, s_k}). \label{eq:ab}
\end{align}
For any ray that does not intersect with boxes, its color and density are set to $0$ and $-\infty$, respectively.
So that the foreground object map $\mathbf{F}$ can be formulated as:
\begin{align}
\mathbf{F}_j =
\begin{cases} 
\mathbf{f}(\mathbf{r}_j),  &\mbox{if } \ \exists \ m \in \mathbf{M}_{:, j}\mbox{, } m \mbox{ is true},\\
0, & \mbox{else}.
\end{cases}
\end{align}
Since the background points are sampled at a fixed depth, we can directly adopt \cref{eq:bg} to evaluate background points in the global space without sorting. 
And the background map $\mathbf{N}$ can also be obtained by volume rendering similar to \cref{eq:integral}.
Finally, $\mathbf{F}$ and $\mathbf{N}$ are alpha-blended into the final image $\mathbf{I}_n$ with alpha extracted from \cref{eq:ab}:
\begin{align}
    \mathbf{I}_\mathrm{n} = \mathbf{F} + 
    \prod_{k=1}^{n_j\mathrm{N}_d}(1-\alpha_{j, k}) \odot \mathbf{N}.
\end{align}
Although our rendering pipeline efficiently composites multiple radiance fields, it still suffers from slow performance when rendering high-resolution images.
To mitigate this issue, we render a high-dimensional feature map instead of a $3$-channel color in a smaller resolution, followed by a StyleGAN2-like architecture that upsamples the feature map to the target resolution.

\subsection{Local \& Global Discrimination} \label{sec:method:discrimination}

Like other GAN-based approaches, discriminators play a crucial role in training.
Previous attempts for 3D-aware scene synthesis\cite{giraffe, gsn} only adopt scene-level discriminators to critique between rendered scenes and real captures. 
However, such a scene discriminator pays more attention to the global coherence of the whole scene, weakening the supervision for individual objects.
Given that each object, especially those far from the camera, occupies a small portion of the rendered frame, the scene discriminator provides weak learning signal to its radiance field, leading to inadequate training and poor object quality.   
Besides, the scene discriminator shows only minimal capability in disentangling objects and background, allowing the background generator $\mathrm{G_{bg}}$ to overfit the whole scene easily.

Similar to ~\cite{DetailMeMore}, we propose to add an extra object discriminator for local discrimination, leading to better object-level supervision.
Sepcifically, with the 3D layout $\mathbf{B}_i$ spatially disentangling different objects, we project them into 2D space as $\mathbf{B}_i^{2D}$ to extract object patches $\mathcal{P}_{\mathbf{I}} = \{\mathbf{P}_i | \mathbf{P}_i = \mathrm{crop}(\mathbf{I}, \mathbf{B}_i^{2D})\}$ from synthesized and real scenes images with simple cropping.
The object patches are fed into the object discriminator after being scaled to a uniform size.
We find that it significantly helps synthesize realistic objects and benefits the disentanglement between objects and the background.
More details about our object discrimination are included in the \textit{Supplementary Materials}.

\subsection{Training and Inference} \label{sec:method:training}

\noindent\textbf{Training Objectives}.
The whole generation process is formulated as $\mathbf{I}_f=\mathrm{G}(\mathcal{B}, \mathcal{Z}, \xi)$, where the generator $\mathrm{G}(\cdot)$ receives a layout $\mathcal{B}$, a latent code set $\mathcal{Z}$ independently sampled from distribution $\mathcal{N}(0, 1)$ to control objects, and a camera pose $\xi$ sampled from a prior distribution $p_\xi$ to synthesize the image $\mathbf{I}_f$.
During training, $\mathcal{B}$, $\mathcal{Z}$, $\xi$ are randomly sampled,
and the real image $\mathbf{I}_r$ is sampled from the dataset.
Besides the generator, we employ the scene discriminator $\mathrm{D}_{s}(\cdot)$ to guarantee the global coherence of the rendering and the object discriminator $\mathrm{D_{obj}}(\cdot)$ on individual objects for local discrimination.
Generators and discriminators are jointly trained as: 
\begin{align}
    \min\mathcal{L}_G &= \mathbb{E}[f(-\mathrm{D}_s(\mathbf{I}_f))] + \lambda_1 \mathbb{E}[f(-\mathrm{D_{obj}}(\mathcal{P}_{\mathbf{I}_{f}}))],   \\
    \min\mathcal{L}_D &= \mathbb{E}[f(-\mathrm{D}_s(\mathbf{I}_r))] +                                     \mathbb{E}[f(\mathrm{D}_s(\mathbf{I}_f))]   \label{eq:d_loss}     
                        \\ \nonumber
                        & +\lambda_1 (\mathbb{E}[f(-\mathrm{D_{obj}}(\mathcal{P}_{\mathbf{I}_{r}})]) +\mathbb{E}[f(\mathrm{D_{obj}}(\mathcal{P}_{\mathbf{I}_{f}})))]
                        \\ \nonumber
                         &+\lambda_2 ||\nabla_{\mathbf{I}_{r}}\mathrm{D}_s(\mathbf{I}_{r})||_2^2 + \lambda_3 \nabla_{\mathcal{P}_{\mathbf{I}_{r}}}\mathrm{D_{obj}}(\mathcal{P}_{\mathbf{I}_{r}})||_2^2) , 
\end{align}
where $f(t) = \log(1+\exp(t))$ is the softplus function, and $\mathcal{P}_{\mathbf{I}_{r}}$ and $\mathcal{P}_{\mathbf{I}_{f}}$ are the extracted object patches of synthesized image $\mathbf{I}_{f}$ and real image $\mathbf{I}_{r}$, respectively.
$\lambda_1$ stands for the loss weight of the object discriminator. 
The last two terms in \cref{eq:d_loss} are the gradient penalty regularizers of both discriminators, with $\lambda_2$ and $\lambda_3$ denoting their weights.

\noindent\textbf{Inference}. 
Besides high-quality scene generation, our method naturally supports object editing by manipulating the layout prior as shown in \cref{fig:pipeline}. Various applications are shown in \cref{sec:exp:editing}.
In particular, ray marching at a small resolution ($64$) may cause aliasing especially when moving the objects.
We adopt supersampling anti-aliasing (SSAA)~\cite{ggp} to perform ray marching at a temporary higher resolution ($128$) and downsample the feature map to the original resolution before the upsampler.
This strategy is used only for object synthesis, and we do not change the background resolution during inference.

\section{Experiments}\label{sec:exp}

\begin{figure*}[t]
\begin{center}
\includegraphics[width=0.9\linewidth]{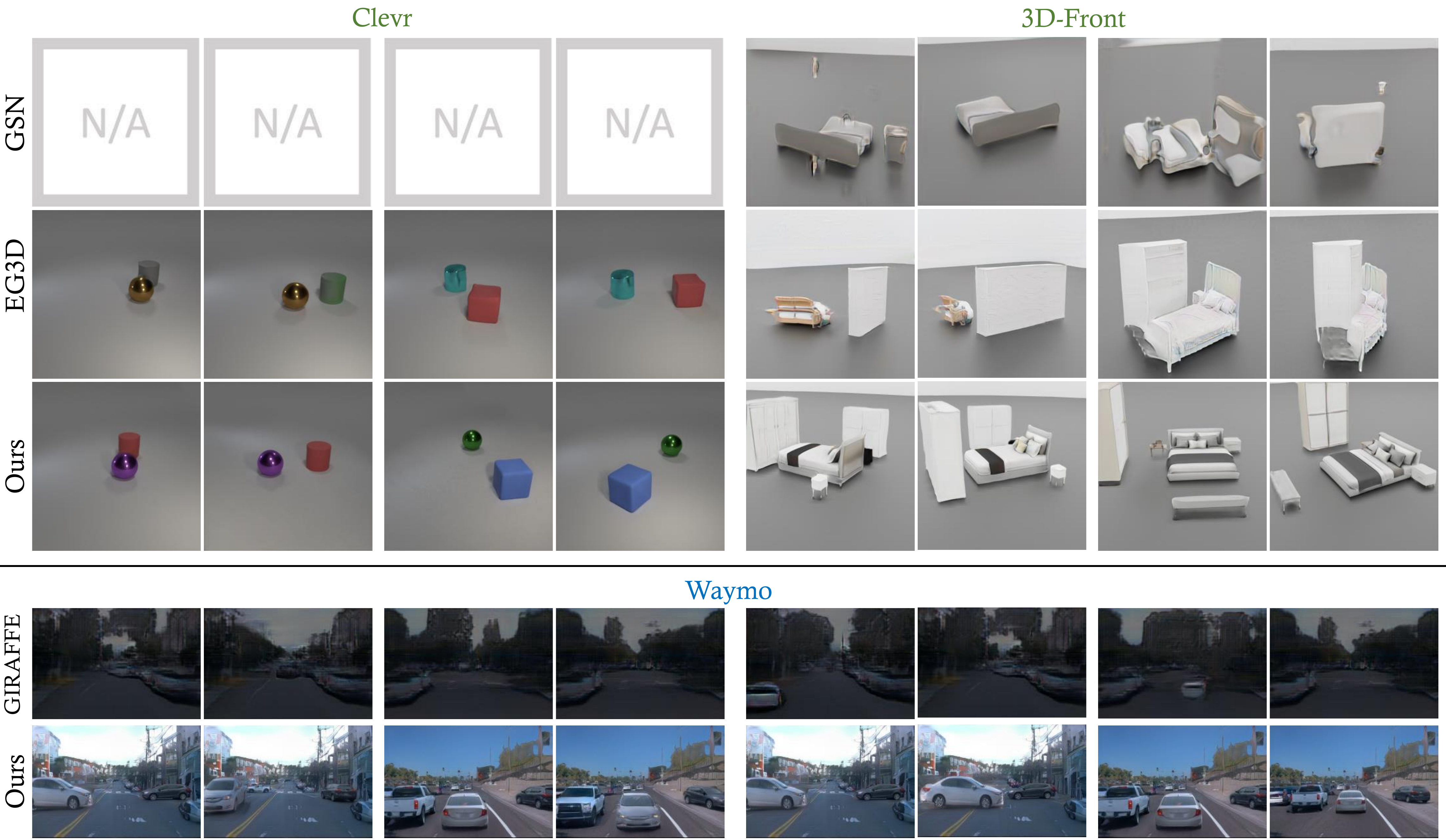}
\end{center}
\vspace{-18pt}
\caption{\textbf{Qualitative comparison between \method and baselines}. {\color{color1} Explicit camera rotation} is evaluated on \clevr  and \front. {\color{color2} Object rotation} (left) and {\color{color2} object translation} (right) are evaluated on \waymo. All images are in $256\times256$ resolution. 
}

\label{fig:main-results}
\vspace{-8pt}
\end{figure*}

\subsection{Settings}
\noindent{\textbf{Datasets}}.
We evaluate \method on three multi-object scene datasets, including \clevr~\cite{clevr}, \front~\cite{3dfront, 3dfuture}, and \waymo~\cite{waymo}.
\clevr \ is a diagnostic multi-object dataset. We use the official script~\cite{clevr} to render scenes with 2 and random primitives.
Our \clevr dataset consists of $80K$ samples in $256\times256$ resolution.
\front is an indoor scene dataset, containing a collection of $6.8K$ houses with $140K$ rooms. 
We obtain $4K$ bedrooms after filtering out rooms with uncommon arrangements or unnatural sizes and use \href{https://github.com/DLR-RM/BlenderProc}{BlenderProc} to render $20$ images per room from random camera positions, resulting in a total of $80K$ images.
\waymo is a large-scale autonomous driving dataset with $1K$ video sequences of outdoor scenes.
Six images are provided for each frame, and we only keep the front view.
We also apply heuristic rules to filter out small and noisy cars and collect a subset of $70K$ images. 
Because the width is always larger than height on \waymo, we adopt the black padding to make images square, similar with StyleGAN2~\cite{stylegan2}.
More details about data preprocessing and rendering are included in \supp.

\noindent{\textbf{Baselines}}.
We compare with both 2D and 3D GANs.
For 2D, we compare with StyleGAN2~\cite{stylegan2} on image quality.
As for 3D, we compare with EpiGRAF~\cite{epigraf}, VolumeGAN~\cite{volumegan}, and EG-3D~\cite{eg3d} on object generation, and GIRAFFE~\cite{giraffe}, GSN~\cite{gsn} on scene generation.
We use the baseline models either released along with their papers or official implementations to train on our data.%
\footnote{We fail to train GSN on \clevr and \waymo with the \href{https://github.com/apple/ml-gsn}{official implementation}, hence we do not report the quantitative results.}

\noindent{\textbf{Implementation Details}}.
We use the same architecture and parameters of the mapping network from StyleGAN2~\cite{stylegan2}.
For object generator $\mathrm{G_{obj}}(\cdot)$ and background generator $\mathrm{G_{bg}}(\cdot)$, we use $8$ and $4$ Modulated Fully-Connected layers (ModFCs) with $256$ and $128$ channels, respectively.
Ray casting is performed on $64\times64$ and the feature map is rendered to image with neural renderer.
The progressive training strategy from PG-GAN~\cite{pggan} is adopted for better image quality and multi-view consistency.
Discriminators $\mathrm{D}_{s}(\cdot)$ and  $\mathrm{D_{obj}}(\cdot)$ both share the similar architecture of StyleGAN2 but with only half channels.
Practically, the resolution of $\mathrm{D_{obj}}(\cdot)$ is always $1/2$ on \waymo or $1/4$ on \clevr and \front of $\mathrm{D}_{s}(\cdot)$. 
All our models are trained on $8\times$ V100/A100 GPUs with a batch size of $64$. 
$\lambda_1$ is set to $1$ to balance object and scene discriminators.
$\lambda_2$ and $\lambda_3$ are set to $1$ to maintain training stability.
Unless specified, other hyperparamters are same as StyleGAN2. 
More details about network architecture and training can be found in \supp.

\subsection{Main Results}

\setlength{\tabcolsep}{8pt}
\begin{table*}[t]
\center
\caption{ \textbf{Quantitative comparisons on different datasets.} FID, KID ($\times 10^3$) are reported as the evaluation metrics. TR. and INF. denote training and inference costs, evaluated in \textit{V100 days} and \textit{ms/image} (single V100 over $1K$ samples), respectively. Note that we highlight the best results among 3D-aware models.} \label{tab:main-results}
\vspace{-20pt}
\begin{tabular}{lccccccccc}
\multicolumn{8}{c}{} \\
\toprule
\multirow{2}{*}{Model} & \multicolumn{4}{c}{\clevr} & \multicolumn{2}{c}{\front} & \multicolumn{2}{c}{\waymo} \\
\cmidrule(lr){2-5} 
\cmidrule(lr){6-7} 
\cmidrule(lr){8-9} 
& FID $\downarrow$ & KID $\downarrow$  &  TR. $\downarrow$ & INF. $\downarrow$ &  FID $\downarrow$ & KID $\downarrow$  & FID $\downarrow$ & KID $\downarrow$ & \\
\midrule
StyleGAN2~\cite{stylegan2}  & 4.5 & 3.0  &  13.3 & 44 & 12.5 & 4.3  & 15.1 & 8.3  \\ 
\midrule
EpiGRAF~\cite{epigraf} & 10.4 & 8.3 & 16.0 & 114 & 107.2 &  102.3 & 27.0 & 26.1  \\
VolumeGAN~\cite{volumegan} & 7.5 & 5.1  & 15.2 & 90 & 52.7 & 38.7  & 29.9 & 18.2 &\\
EG3D~\cite{eg3d} & 4.1 & 12.7  & 25.8 & \textbf{55} & 19.7 & 13.5  &26.0 & 45.4 \\
\midrule
GIRAFFE~\cite{giraffe} & 78.5 & 61.5 & \textbf{5.2}& 62 & 56.5  & 46.8  & 175.7  & 212.1  \\
GSN~\cite{gsn} & $-$ & $-$  & $-$ & $-$ &130.7 & 87.5  & $-$ & $-$ \\
\midrule
\rowcolor{LightGray}
DisCoScene  & \textbf{3.5} & \textbf{2.1}  & 18.1 & 95 & \textbf{13.8} & \textbf{7.4}  & \textbf{16.0}  & \textbf{8.4}  \\
\bottomrule
\end{tabular}
\vspace{-7pt}
\end{table*}

\noindent{\textbf{Qualitative Comparison}}.
\cref{fig:main-results} presents the synthesized images in a resolution of $256\times256$ of our method and baselines on all the datasets.
We compare our method on explicit camera control and object editing with baselines.

GSN and EG3D, with a single radiance field, can manipulate the global camera of the synthesized images.
GSN highly depends on the training camera trajectories.
Thus in our setting where the camera positions are randomly sampled, it suffers from the poor image quality and multi-view consistency.
As for EG3D, although it converges on the datasets, the object fidelity are lower than our method.
On \clevr  with a narrow camera distribution, the results of EG3D are inconsistent.
In the first example, the color of the cylinder changes from gray to green across different views. 
Meanwhile, our method learns better 3D structure of the objects and achieves better camera control.
On the challenging \waymo dataset, it is difficult to encode huge street scenes within a single generator, thus we train GIRAFFE and our \method in the camera space to evaluate object editing.
GIRAFFE struggles to generate realistic results and, while manipulating objects, their geometry and appearance are not preserved well.
Our approach is capable of handling these complicated scenarios with good variations.
Wherever the object is placed and regardless of how the rotation is carried out, the synthesized objects are substantially better and more consistent than GIRAFFE.
It demonstrates the effectiveness of our spatially disentangle radiance fields built upon the layout prior.

\noindent{\textbf{Quantitative Comparison}}.
\cref{tab:main-results} reports the quantitative metrics on the quality of results, including FID~\cite{fid} and KID~\cite{kid}.
All metrics are calculated between $50K$ generated samples and all real images.
\method consistently outperforms baselines with significant improvement on all datasets.
Besides, training cost in \textit{V100 days} and testing cost in \textit{ms/image} (on a single V100 over $1K$ samples) are also included to reflect the efficiency of our model.
Note that the inference cost of 3D-aware models is evaluated on generating radiance fields rather than images.
In such a case, EG3D and EpiGRAF are not fast as excepted due to the heavy computation on tri-planes.
With comparable training and testing cost, it even achieves similar level of image quality with state-of-the-art 2D GAN baselines, \eg, StyleGAN2~\cite{stylegan2}, while allowing for explicit camera control and object editing that are otherwise challenging.

\definecolor{origin}{HTML}{F4B183}
\begin{figure*}[t]
\begin{center}
\includegraphics[width=1\textwidth]{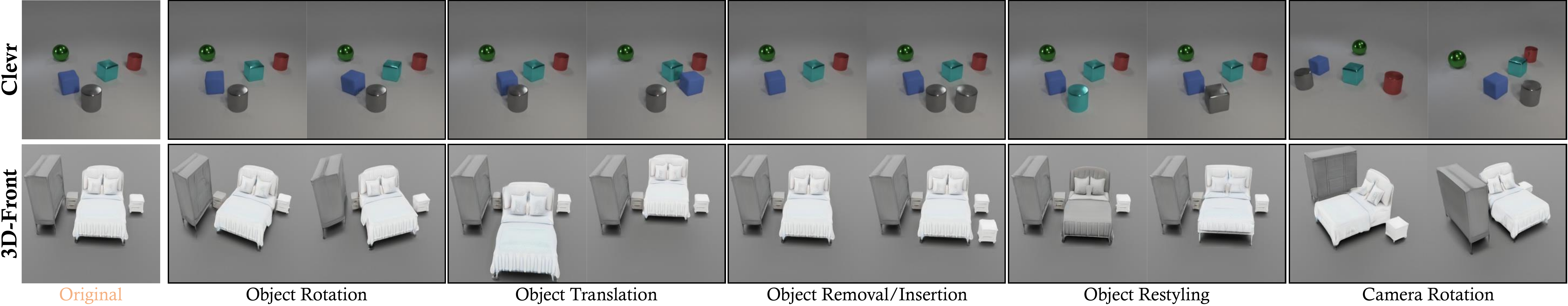}
\includegraphics[width=1\textwidth]{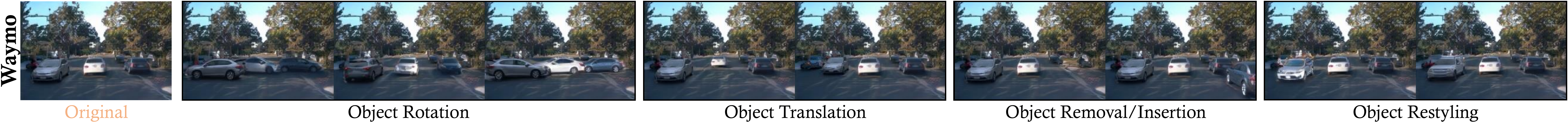}
\end{center}
\vspace{-20pt}
\caption{\textbf{Controllable scene synthesis in $256\times256$ resolution.}  We perform versatile user control of the global camera and scene objects, such as rearrangement, removal, insertion, and restyling. Editings are performed on the {\color{origin}original generations} (the left one of each row).}
\label{fig:object-edting}
\vspace{-10pt}
\end{figure*}

\subsection{Controllable Scene Generation} \label{sec:exp:editing}
The layout prior in our model enables versatile user controls of scene objects. 
In what follows, we evaluate the flexibility and effectiveness of our model through various 3D manipulation applications in different datasets. Examples are shown in \cref{fig:object-edting} and more results can be found in \supp.

\noindent{\textbf{Rearranging Objects}}. 
We can transform bounding boxes $\mathbf{B}$ to rearrange (rotation and translation) the objects in the scenes without affecting their appearance.
Transforming shapes in \clevr, furniture in \front, and cars in \waymo \ all show consistent results.
In particular, rotating symmetric shapes (\ie, spheres and cylinders) in \clevr  shows little changes, suggesting desired multi-view consistency.
Our model can properly handle mutual occlusion.
Take the blue cube from \clevr \ as example ($1$-st row of \cref{fig:object-edting}), our model can produce new occlusions between it and the grey cylinder and generate high-quality renderings.

\noindent{\textbf{Removing and Cloning Objects}}.
Users can update the layout by removing or cloning bounding boxes.
Our method seamlessly removes objects with the background inpainted realistically, even without training on any pure background, including the challenging dataset of \waymo($3$-rd row of \cref{fig:object-edting}).
Object cloning is also naturally supported, by copying and pasting a box to a new location in the layout.

\noindent{\textbf{Restyling Objects}}.
Although appearance and shape are not explicitly modeled by the latent code, we can reuse the encoded hierarchical knowledge to perform object restyling.
Like~\cite{ghfeat, stylenerf, interfacegan}, we arbitrarily sample latent codes and perform style-mixing on different layers to achieve independent control over appearance and shape.
\cref{fig:object-edting} presents the restyling results on certain objects, \ie, the front cylinder in \clevr, the bed in \front, and the left car in \waymo.

\noindent{\textbf{Camera Movement}}.
Explicit camera control is also permitted.
Even for \clevr  that is trained on very limited camera ranges, we can rotate the camera up to an extreme side view.
Our model also produces consistent results when rotating the camera on \front ($2$-nd row of \cref{fig:object-edting}).

\subsection{Ablation Study}

\begin{figure*}[t]
    \centering
    \hspace{-3mm}
    \subfloat[Spatial condition \label{fig:spatial-cond}]{
        \includegraphics[width=0.338\textwidth]{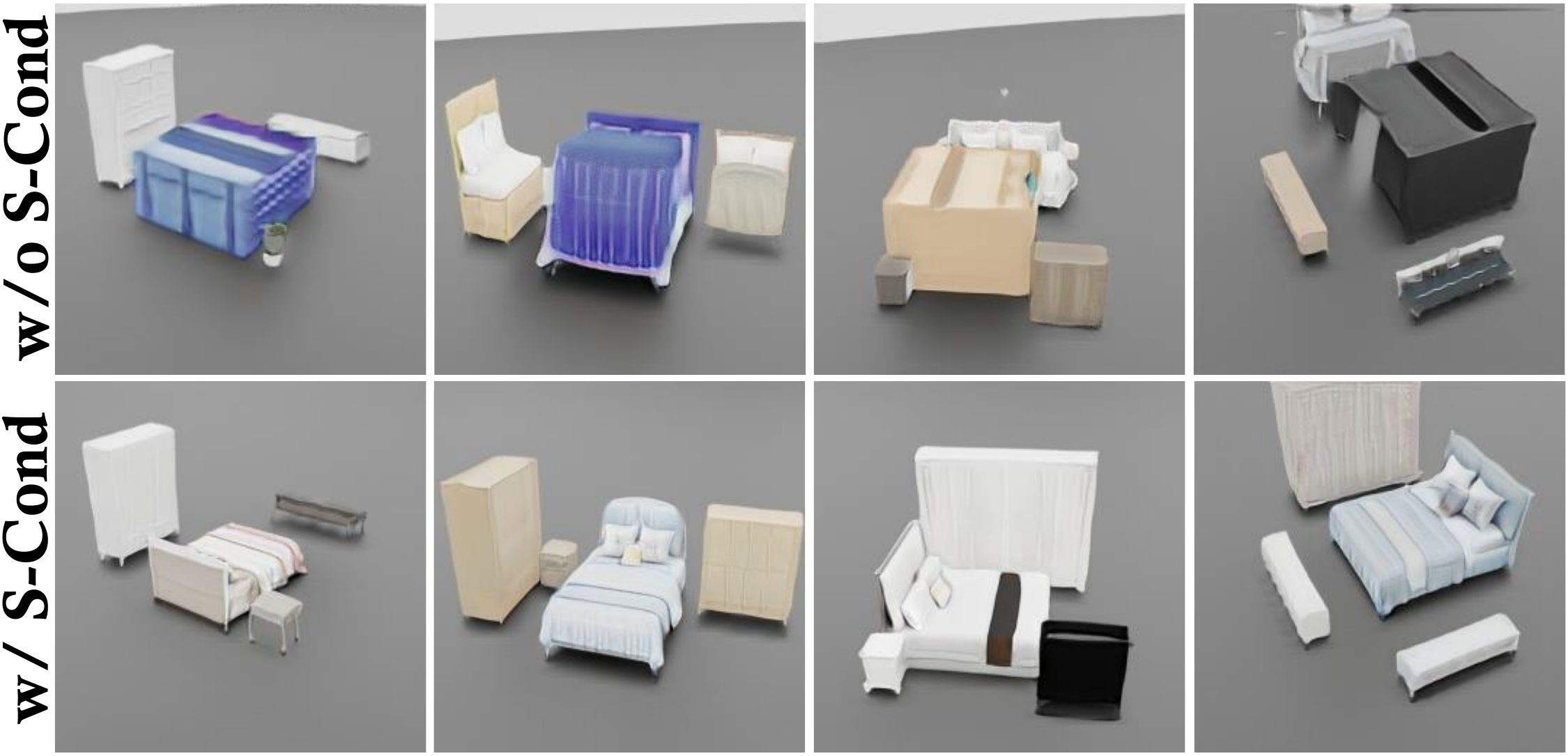}
    }\hspace{-2mm}
    \subfloat[Supersampling anti-aliasing \label{fig:anti-aliasing}]{
        \includegraphics[width=0.32\textwidth]{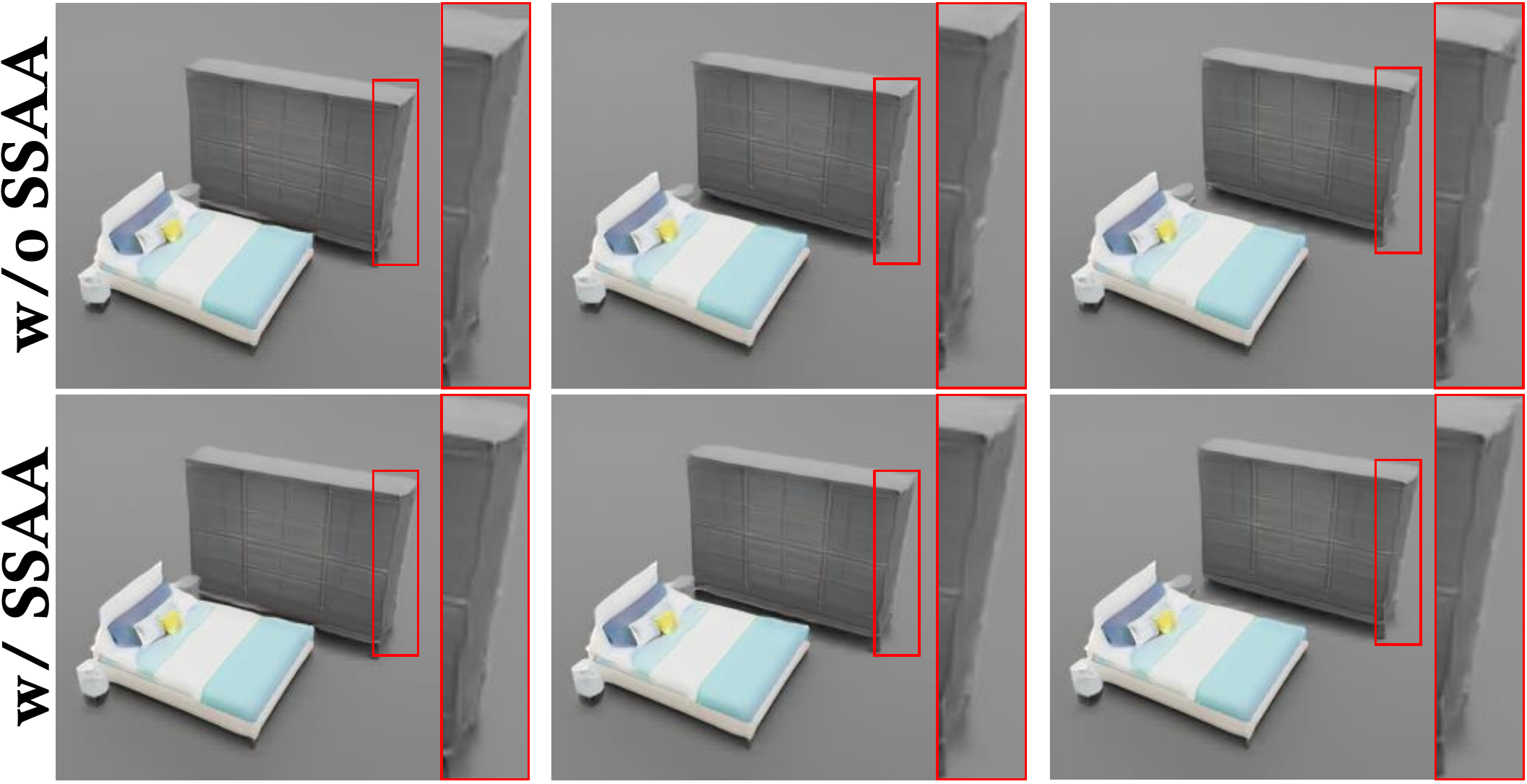}
    }\hspace{-2mm}
    \subfloat[Upsampler \label{fig:upsampler}]{
        \includegraphics[width=0.338\textwidth]{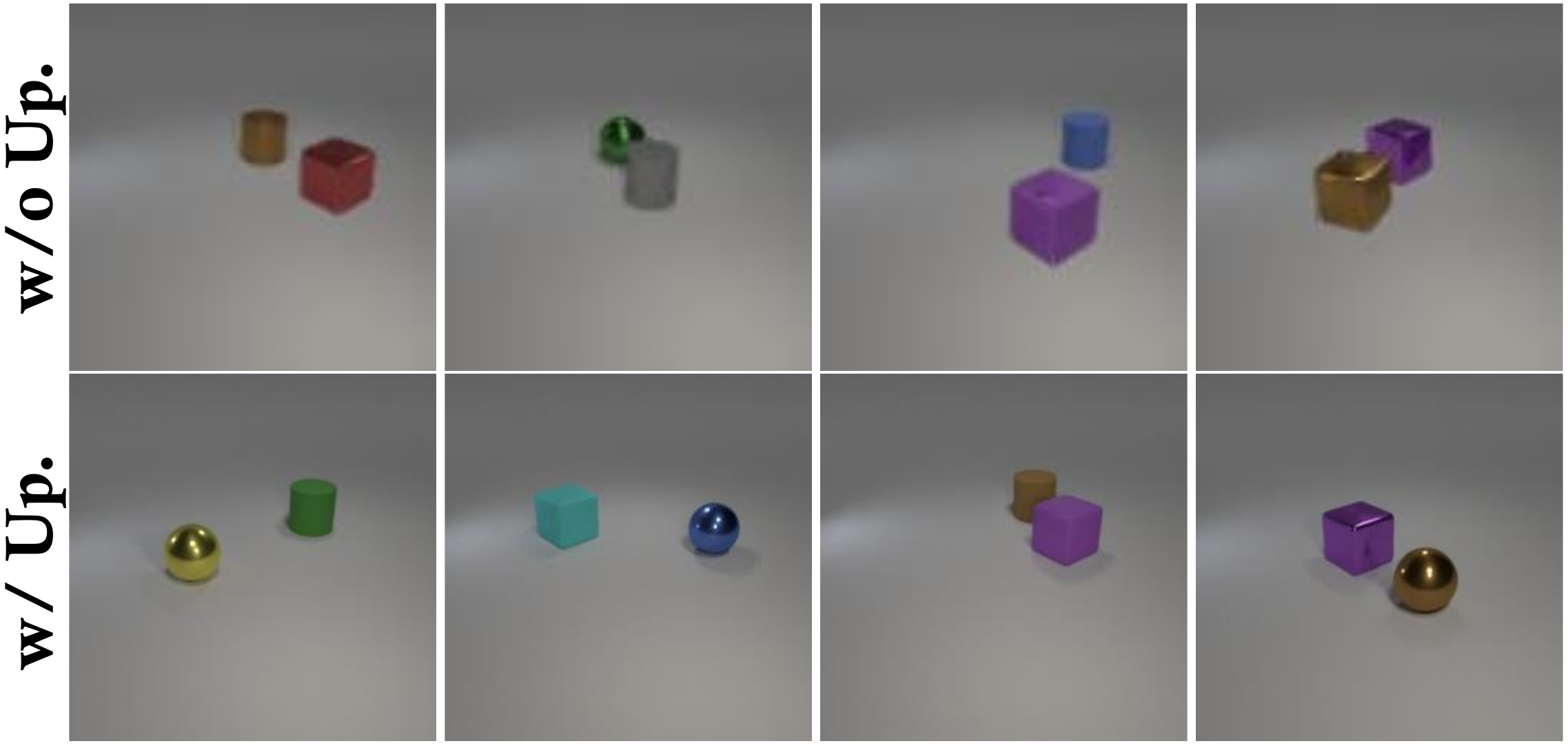}
    }
\vspace{-10pt}
\caption{\textbf{Qualitative comparison for ablations} on spatial condition (S-Cond), supersampling anti-alising (SSAA), and upsampler (Up.).}
\vspace{-15pt}
\end{figure*}

We ablate main components of our approach to better understand their individual contributions.
In addition to the FID score that measures the quality of the entire image, we also provide another metric FID$_\textrm{obj}$ to measure the quality of individual objects.
Specifically, we use the projected 2D boxes to crop objects from the synthesized images and then perform FID evaluation against the ones from real images.

\definecolor{azure}{rgb}{0.0, 0.5, 1.0}
\setlength{\tabcolsep}{5pt}
\begin{table}[t]
\center
\caption{\textbf{Ablation analysis} of object discriminator ($\mathrm{D_{obj}}$).}  \label{tab:objectd}
\vspace{-8pt}
\begin{tabular}{ll|ccc}
\toprule
   & & \clevr & \front & \waymo \\
\hline
\multirow{2}{*}{FID} & \textit{w/o} $\mathrm{D_{obj}}$ & 5.0 & 18.6 & 19.5\\
                        & \textit{w/} $\mathrm{D_{obj}}$ & \textbf{3.5} & \textbf{13.8}& \textbf{16.0}\\
\midrule
\multirow{2}{*}{FID$_\textrm{obj}$} & \textit{w/o} $\mathrm{D_{obj}}$ & 19.1 & 33.7 & 95.1 \\
                        & \textit{w/} $\mathrm{D_{obj}}$ & \textbf{5.6}& \textbf{19.5} & \textbf{16.3}\\
\bottomrule
\end{tabular}
\vspace{-10pt}
\end{table}

\setlength{\tabcolsep}{3pt}
\begin{table}[t]
\center
\caption{\textbf{Ablation analysis} of spatial condition (S-Cond).} \label{table:scond}
\vspace{-20pt}
\begin{tabular}{lcc|cc}
\multicolumn{5}{c}{} \\
\toprule
\multirow{2}{*}{} & \multicolumn{2}{c|}{FID} & \multicolumn{2}{c}{FID$_\textrm{obj}$} \\ 
\cline{2-5}
&  \textit{w/} \footnotesize{S-Cond} & \textit{w/o} \footnotesize{S-Cond} &  \textit{w/} \footnotesize{S-Cond} & \textit{w/o} \footnotesize{S-Cond} \\
\hline
\front & \textbf{13.8} & 15.2 & \textbf{19.5} & 23.2\\
\bottomrule
\end{tabular}
\vspace{-10pt}
\end{table}

\begin{figure}[t]
\begin{center}
\includegraphics[width=1\linewidth]{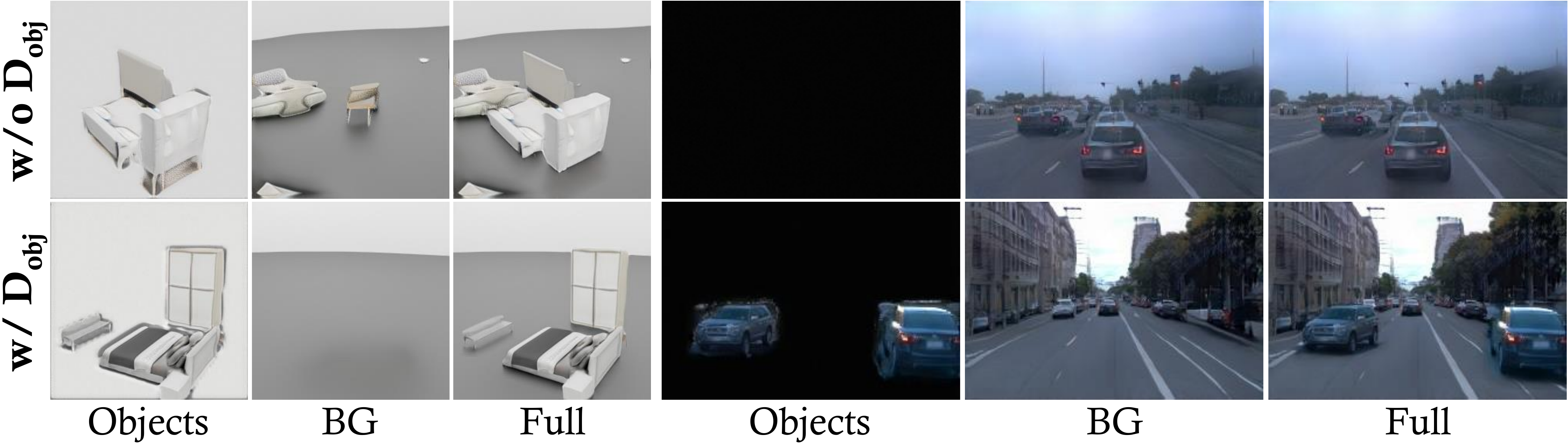}
\end{center}
\vspace{-15pt}
\caption{\textbf{Ablation on scene disentanglement}. We independently infer objects and background to show the quality of scene disentanglement with regard to object discriminator $\mathrm{D_{obj}}$.}
\label{fig:fg-bg-dis}
\vspace{-10pt}
\end{figure}

\noindent{\textbf{Object Discriminator}}.
The object discriminator $\mathrm{D_{obj}}$ plays a crucial role in synthesizing realistic objects, as evaluted in \cref{tab:objectd}.
Obviously, the object fidelity is significantly improved across all datasets with $\mathrm{D_{obj}}$.
Also, the quality of the whole scene generation is improved as well, contributed by better objects.
\cref{fig:fg-bg-dis} visually shows that our method can successfully disentangle objects from the background with the help of object discriminator.
Although the baseline model is able to disentangle objects on \front from simple background to certain extent, the background suffers from the entanglement with objects, resulting in obvious artifacts as well as illogical layout.
On more challenging datasets like \waymo, the complex backgrounds make the disentanglement even more difficult, so that the background model easily overfits the whole scene as a single radiance field.
Thanks to the object discriminator, our full model benefits from object supervision, leading to better disentanglement, even without seeing a pure background image.

\noindent{\textbf{Spatial Condition}}.
To analyze how spatial condition (S-Cond) affects the quality of generation, we compare results with models trained with and without S-Cond on \front (\cref{fig:spatial-cond}).
For example, our full model consistently infers beds at the center of rooms, while the baseline predicts random items like tables or nightstands that rarely appear in the middle of bedrooms.
These results demonstrate that spatial condition can assist the generator with appropriate semantics from simple layout priors.
Note that this correlation between spatial configurations and object semantics is automatically emerged without any supervision.
We also numerically compare the image quality on these two models in \cref{table:scond}, which shows that S-Cond also achieves better image quality at both scene- and object-level, because more proper semantics are more in line with the native distribution of real images.

\noindent{\textbf{Supersampling Anti-Aliasing}}. 
We adopt a simple super-sampling (SSAA) strategy to reduce edge aliasing by sampling more points during inference (\cref{sec:method:training}). 
Thanks to our efficient object point sampling, doubling the resolution of foreground points keeps a similar inference speed ($105$ \textit{ms/image}), comparable with original speed ($95$ \textit{ms/image}). 
Results with different sampling points are shown in \cref{fig:anti-aliasing}. 
Taking the right boundary of the cabinet as an example (see the zoom-in insets for better visualization), when the cabinet is moved, SSAA achieves more consistent boundary compared with the jaggy one in the baseline. 

\noindent{\textbf{Neural Renderer for Shadow}}.
We adopt the StyleGAN2-like neural renderer to boost the rendering efficiency (\cref{sec:method:rendering}).
Besides the low computational cost, the added capacity of the neural renderer also brings better implicit modeling of realistic lighting effects such as shadowing.
Therefore, without handling the shadowing effect in our rendering pipeline, our model can still synthesize high-quality shaodws on datasets such as \clevr \ (\cref{fig:upsampler}).
This is because the large receptive field brought by $3\times3$ convolutions and upsampler blocks make the neural renderer be aware of the object locations and progressively add shadows to the low resolution features rendered from radiance fields.

\vspace{-5pt}
\section{Discussion and Conclusion}

\begin{figure}[t]
\center
\includegraphics[width=1\linewidth]{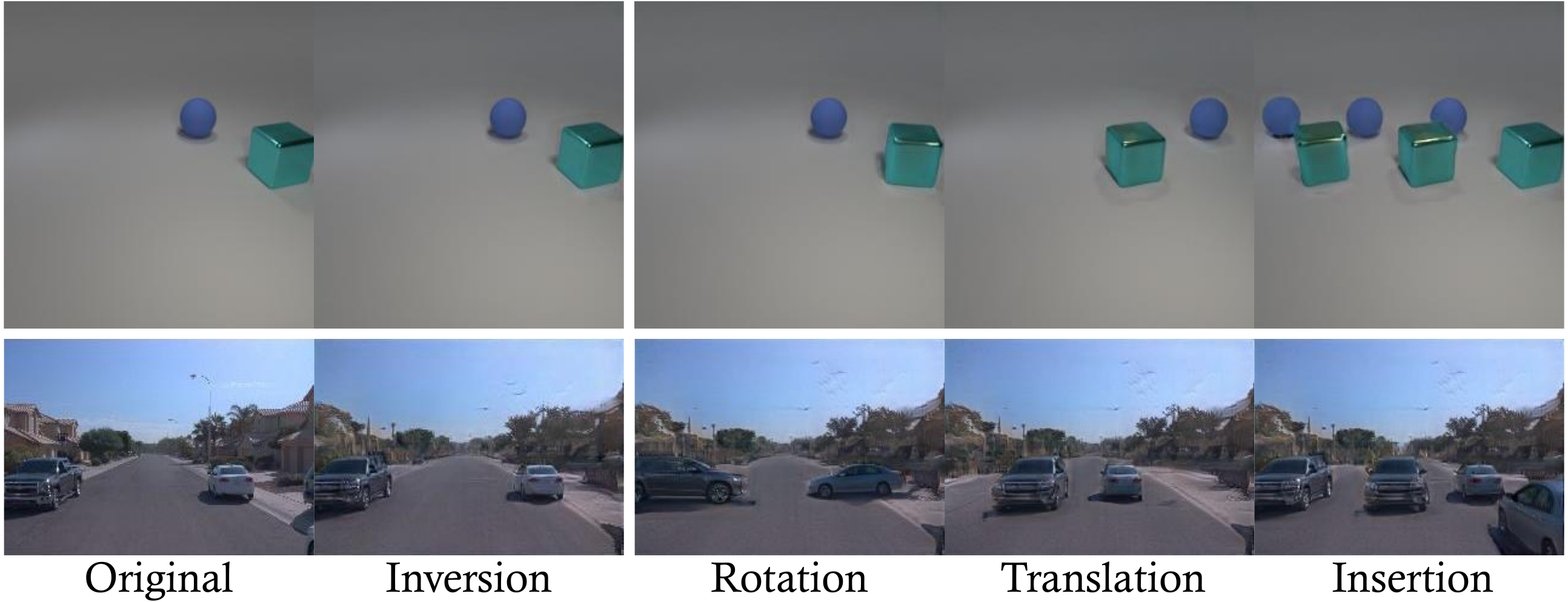}
\vspace{-15pt}
\caption{\textbf{Real image inversion and editing}.}
\label{fig:inversion}
\vspace{-10pt}
\end{figure}

\noindent{\textbf{Real Image Editing}}. 
\cref{fig:inversion} shows that it is possible to embed a real image into the latent space of our pretrained model using pivotal tuning inversion (PTI)~\cite{roich2022pivotal}.
Besides reconstruction, all object manipulation operations are supported to edit the image.
As one of the very first steps towards 3D scene editing from a single image, we believe that our method proves a promising venue and can inspire future research efforts along this direction.

\vspace{2pt}
\noindent{\textbf{Limitations and Future Work}}. 
Our model requires the abstract layout prior as the input.
For in-the-wild datasets, we need monocular 3D object detector~\cite{Wang_2021_ICCV} to infer pseudo layouts.
While existing approaches attempt to learn the layout in an end-to-end manner, they struggle to generalize to complex scenes consisting of multiple objects.
So it would be interesting to explore 3D layout estimation for complex scenes and combine with our approach end-to-end.
Also, although our work shows significant improvement over existing 3D-aware scene generators, it is still challenging to learn on the street scenes in the global space due to the limited model capacity. 
Large-scale NeRFs~\cite{xiangli2022bungeenerf, tancik2022block} might be one potention solutions.

\vspace{2pt}
\noindent{\textbf{Conclusion.}}
This work presents \method, a method for controllable 3D-aware scene synthesis on challenging datasets. By taking spatially disentangled radiance fields as the representation based on a very abstract layout prior, our method is able to generate high-fidelity scene images and allows for versatile object-level editing.

\noindent\textbf{Acknowledgements}. We thank Jiatao Gu, Willi Menapace, Jian Ren, Panos Achlioptas, Tai Wang, and Zian Wang for fruitful discussions and comments about this work.

{\small
\bibliographystyle{ieee_fullname}
\bibliography{ref}
}

\newpage

\section*{A1. Implementation Details of \method} \label{sec:ours}

\noindent{\textbf{Background with NeRF++~\cite{nerf++}}}.
The outdoor datasets, \ie \waymo, have unbounded backgrounds.
It is insufficient to model the whole scene in the image within a fixed bounding box.
Therefore, we inherit the inverse parametrization of NeRF++ to model the background in \waymo:
\begin{align}
    \vx = (x/r, y/r, z/r),
\end{align}
where $r = ||\vx||^2$.
The background points are uniformly sampled in an inverse depth range of $[1/R, 0)$ where $R = 2.0$ denotes the starting depth of the background.

\noindent{\textbf{Constant Latents for Upsampler}}.
We adopt similar architecture and parameters of the synthesis network from StyleGAN2~\cite{stylegan2} as the upsampler for the rendered 2D feature map.
Note that since our model handles multiple radiance fields, different spatial locations of the convolution feature maps should be modulated by different $\vw$ codes belonging to specific objects, making it costly to upsample the feature map.
Thus we disable the spatial-aware modulation by setting $\vw$ as a constant tensor with value $1$, which significantly reduces the computation overhead.

\begin{table*}[t]
    \centering
    \caption{
        \textbf{Training configurations} regarding different datasets for VolumeGAN and EpiGRAF.
    }
    \label{tab:config2}
    \vspace{-5pt}
    \centering
    \resizebox{\textwidth}{!}{
    \begin{tabular}{l|ccccccccc}
    \toprule
            Datasets & FOV & Radius & Range$_{\textrm{depth}}$ & $\#$Steps & Range$_h$ & Range$_v$ & Sample\_Dist & $\lambda$ \\ \hline
     \clevr  & 12.0 & 1.0& $[0.8, 1.2]$ & 24 & $[\pi/2-0.5, \pi/2+0.5]$ & $[\pi/4-0.15, \pi/4+0.15]$ & Uniform & 1 \\
     \front  & 12.8 & 1.0 & $[0.7, 1.3]$ & 24 & $[0, 2\pi]$ & $[3\pi/8-0.2, 3\pi/8+0.2]$ & Uniform & 1   \\
    \waymo     & 12.0 & 1.0 &$[0.7, 1.3]$ & 24 & $[\pi/2-0.5, \pi/2+0.5]$ & $[\pi/2-0.15, \pi/2+0.15]$ & Uniform & 1  \\
    \bottomrule
    \end{tabular}
    }
\end{table*}

\begin{table*}[t]
    \centering
    \caption{
        \textbf{Training configurations} regarding different datasets for EG3D.
    }
    \label{tab:config1}
    \vspace{-5pt}
    \centering
    \resizebox{\textwidth}{!}{
    \begin{tabular}{l|ccccccccc}
    \toprule
        Datasets & FOV & Radius & Range$_{\textrm{depth}}$ & $\#$Steps & Range$_h$ & Range$_v$ & Sample\_Dist & $\lambda$ \\ \hline
     \clevr  & 18.0 & 1.7& $[0.1, 2.6]$ & 96 & $[\pi/2-0.5, \pi/2+0.5]$ & $[\pi/4-0.15, \pi/4+0.15]$ & Uniform & 2 \\
     \front  & 18.8 & 2.7 & $[2.2, 3.3]$ & 96 & $[0, 2\pi]$ & $[3\pi/8-0.2, 3\pi/8+0.2]$ & Uniform & 2   \\
    \waymo     & 18.0 & 1.7 &$[0.1, 2.6]$ & 96 & $[\pi/2-0.5, \pi/2+0.5]$ & $[\pi/2-0.15, \pi/2+0.15]$ & Uniform & 5  \\
    \bottomrule
    \end{tabular}
    }
\end{table*}

\section*{A2. Implementation Details of Baselines} \label{sec:baseline-details}
Because of the wildly divergent data distribution, the training parameters vary greatly on different datasets.
\cref{tab:config2} and \cref{tab:config1} list the detailed training configurations of different datasets for each baseline. 
\textit{FOV}, \textit{Range$_{depth}$}, and \textit{$\#$Steps} denote the field of view, the depth range, and the number of sampling steps along a camera ray, respectively.
\textit{Range$_{h}$} and \textit{Range$_{v}$} denote the horizontal and vertical angle ranges of the camera pose $\xi$.
\textit{Sample\_Dist} denotes the sampling scheme of the camera pose. 
We only use Gaussian or uniform sampling in our experiments.
$\lambda$ is the loss weight of the gradient penalty.

\noindent\textbf{VolumeGAN~\cite{volumegan}}.
We use the official implementation of VolumeGAN.%
\footnote{\href{https://github.com/genforce/volumegan}{https://github.com/genforce/volumegan}}
We train VolumeGAN with $25K$ images.
The coordinates range of feature volume is adjustable for different datasets.
We adopt the training configuration in \cref{tab:config2} to train VolumeGAN models.

\noindent\textbf{EpiGRAF~\cite{epigraf}}
We use the official implementation of EpiGRAF.%
\footnote{\href{https://github.com/universome/epigraf}{https://github.com/universome/epigraf}}
We inherit the patch-wise training scheme to train EpiGRAF with the same data and camera parameters at the target resolution shown in \cref{tab:config2}.

\noindent\textbf{EG3D~\cite{eg3d}.}
We use the official implementation of EG3D.
\footnote{\href{https://github.com/NVlabs/eg3d}{https://github.com/NVlabs/eg3d}}
Different from VolumeGAN and EpiGRAF, EG3D renders the whole radiance field within a bounding box, so we inherit the larger camera radius than the ones of EpiGRAF and VolumeGAN for training.
Since the original EG3D requires pose annotations for training, we add a pose sampler in it to enable the training on all three datasets as the global annotations are not always available.
We adjust the loss weight of gradient penalty on different datasets to achieve the best performance. 
Hyperparameters used for training are available in \cref{tab:config1}

\noindent\textbf{GSN~\cite{gsn}.}
We use the official GSN implementation.%
\footnote{\href{https://github.com/apple/ml-gsn}{https://github.com/apple/ml-gsn}}.
GSN highly dependents on input camera sequences and we find it very difficult to converge at a narrow camera distribution, \ie \waymo and \clevr.
On \front, we set the length of camera sequence to $1$, and it can converge to some extent.
We don't leverage depth supervision for a fair comparison with our method.

\noindent\textbf{GIRAFFE~\cite{giraffe}.}
We use the official implementation of GIRAFFE.%
\footnote{\href{https://github.com/autonomousvision/giraffe}{https://github.com/autonomousvision/giraffe}}
The number of boxes for training follows the configuration of ours on each dataset.
The bounding box generator of GIRAFFE is tuned specifically for each dataset for a fair comparison.

\section*{A3. Data Preparation} \label{sec:data}
\noindent\textbf{Clevr~\cite{clevr}}.
We use the official script~\cite{clevr} to render scenes with Cube, Cylinder, and Sphere primitives.
The camera position is jittered in a small scale.  
And the dataset is rendered in a $256\times256$ resolution with $80K$ samples.

\noindent\textbf{3D-Front~\cite{3dfront, 3dfuture}}.
We use \href{https://github.com/DLR-RM/BlenderProc}{BlenderProc} to render $20$ images per room in \front.
We move the center of each room to the coordinate origin and then sample the camera positions on the upper sphere between 2$r$ to 3$r$ where $r$ is the diagonal length of room.

\noindent\textbf{Waymo~\cite{waymo}}.
We only keep the front view of \waymo for the model training.
However, there exist lots of occluded and noisy cars in \waymo, we design several heuristic rules to filter it.
Specifically, we require the camera depth of car is less than 40$m$ and the area of cars is larger than 40000 pixels in original image size ($1920\times 1280$).
We then adopt the black padding to make images square and then resize it in to $256\times256$ resolution.

\section*{A5. Efficiency of Rendering Pipeline} \label{sec:efficient}
Na\"ive point sampling solutions where the density and color of spatial points are inferred with multiple object radiance fields can lead to prohibitive computational overhead.
Therefore we propose an efficient rendering pipeline by only focusing on the valid points within the bounding boxes.
\cref{table:efficiency} presents the training cost in \textit{V100 days} and testing cost in \textit{ms/image} (on a single V100 over $1K$ samples) with our efficient rendering and Na\"ive rendering.
Our rendering pipeline can handle multiple objects efficiently, with nearly 1.5 and 2 times faster training and inference speed, respectively. 

\setlength{\tabcolsep}{5pt}
\begin{table}[t]
\center
\caption{\textbf{Ablation analysis} of efficient rendering (ER).} \label{table:efficiency}
\vspace{-20pt}
\begin{tabular}{l|cc|cc}
\multicolumn{5}{c}{} \\
\toprule
\multirow{2}{*}{} & \multicolumn{2}{c|}{TR.$\downarrow$} & \multicolumn{2}{c}{INF.$\downarrow$} \\ 
\cline{2-5}
&  \textit{w/} \footnotesize{ER} & \textit{w/o} \footnotesize{ER} &  \textit{w/} \footnotesize{ER} & \textit{w/o} \footnotesize{ER} \\
\hline
\clevr & \textbf{18.1} & 29.2 & \textbf{95} & 180\\
\front & \textbf{22.3} & 38.1 & \textbf{110} & 330\\
\waymo & \textbf{19.2} & 30.9 & \textbf{100} & 195\\
\bottomrule
\end{tabular}
\vspace{-10pt}
\end{table}

\section*{A6. Additional Results} \label{sec:additional-results}

We include a \href{https://www.youtube.com/watch?v=Fvenkw7yeok}{demo video}, which shows more results of various 3D manipulation applications.
From the video, we can see that our method both achieves good generation quality and enables precise object control.
We also include comparisons with the state-of-the-art methods, \textit{i.e.}, EG3D~\cite{eg3d} and GIRRAFE~\cite{giraffe}, in the video.

\end{document}